\documentclass{article}

\usepackage{graphicx}
\usepackage{amsmath,amssymb}
\usepackage{booktabs}
\usepackage{multirow}
\usepackage[table]{xcolor}
\usepackage{capt-of}
\usepackage{wrapfig}
\usepackage{pgfplots}
\pgfplotsset{compat=1.18}
\usepackage[most]{tcolorbox}
\usepackage{listings}

\newtcblisting{promptbox}[1][]{
  enhanced,
  breakable,
  colback=gray!4,
  colframe=gray!55,
  boxrule=0.4pt,
  arc=1mm,
  left=0.8mm,
  right=0.8mm,
  top=0.6mm,
  bottom=0.6mm,
  boxsep=0.5mm,
  before skip=0.5em,
  after skip=0.5em,
  fonttitle=\bfseries\scriptsize,
  listing only,
  listing options={
    basicstyle=\ttfamily\tiny,
    breaklines=true,
    breakatwhitespace=false,
    columns=fullflexible,
    keepspaces=true,
    showstringspaces=false
  },
  #1
}

\usepackage[preprint]{corl_2026} 

\title{DriveMA: Driving Vision-Language-Action Models with verifiable Meta-Actions}

\author{
  Weicheng Zheng$^{1,2}$ \qquad
  Yixin Huang$^{1,3}$ \qquad
  Qiao Sun$^{1}$ \qquad
  Derun Li$^{1}$ \qquad
  Hang Zhao$^{1,2}$\thanks{Corresponding author: hangzhao@mail.tsinghua.edu.cn} \\
  \\
  $^1$Shanghai Qi Zhi Institute \qquad
  $^2$IIIS Tsinghua University \qquad
  $^3$Tongji University
}

\begin{document}
\maketitle

\vspace{-5pt}
\begin{abstract}
Driving Vision-Language-Action Models (Driving VLAs) aim to use language to improve end-to-end planning, but the language-action gap limits this promise. 
We propose \textbf{DriveMA}, a Driving VLA framework built on verifiable meta-actions, which summarize future ego motion into compact language-domain intentions and can be constructed from expert trajectories with a trajectory-grounded annotation pipeline and can be verified against generated trajectories through rule-based projection.
DriveMA exploits this verifiability with action-centric supervised training and a data-efficient turn-level credit assignment reinforcement learning framework, explicitly aligning high-level decisions with low-level trajectory planning through dense rewards and precise credit assignment. 
DriveMA sets a new state of the art on the Waymo Open Dataset Vision-based E2E Driving, achieving a Rater Feedback Score of \textbf{8.060} with a 2B model and further improving it to \textbf{8.079} with a 4B model; it also obtains competitive closed-loop planning performance on NAVSIM.
These results show that even a simple meta-action interface can achieve state-of-the-art planning when made verifiable and optimized for language-action alignment.
Code, data, and models are available at https://tsinghua-mars-lab.github.io/DriveMA.
\end{abstract}



\section{Introduction}

\begin{figure}[!htbp]
    \centering
    \includegraphics[width=\linewidth]{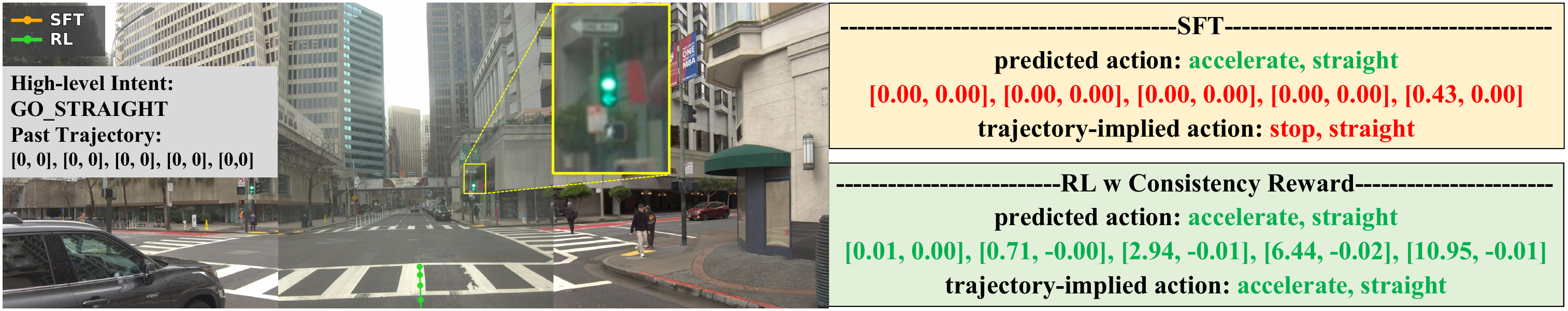}
    \caption{
        \textbf{Motivating example of the language-action gap.}
At a green light with near-static motion history, the SFT-only DriveMA predicts a plausible high-level decision to accelerate, but generates an almost stationary trajectory, revealing a language-action mismatch.
With our RL-based explicit alignment, the trajectory better executes the predicted acceleration decision.
}
    \label{fig:case1}
\end{figure}

Driving Vision-Language-Action Models (Driving VLAs) have recently emerged as a promising paradigm for end-to-end autonomous driving~\cite{shao2024lmdrive,sima2024drivelm,tian2024drivevlm,zhou2025autovla}. 
By introducing language into the perception-to-action pipeline, these models aim to leverage semantic knowledge to improve downstream planning. 
Ideally, language should not merely describe the scene, but should expose high-level driving intentions that guide low-level trajectory generation. 
However, the persistent language-action gap in Driving VLAs means that plausible language-side decisions may not be faithfully executed by predicted trajectories, preventing language from becoming a truly actionable planning interface~\cite{renz2025simlingo,li2025driver1}. Fig.~\ref{fig:case1} illustrates a concrete language-action mismatch: despite predicting acceleration at a green light, an SFT model remains nearly stationary under a near-static motion history.

We argue that bridging this gap requires the intermediate language interface to be \emph{verifiable}. 
A verifiable interface allows high-level language decisions to be supervised from expert behavior, checked against generated trajectories, and optimized through explicit alignment. 
In this paper, we study \emph{meta-action} as a simple and effective instance of such a verifiable language interface. 
A meta-action compactly represents the planning-relevant intent of future ego motion, such as accelerating, decelerating, turning, or changing lanes.
More importantly, trajectories can be projected back into a verifiable action space, making it possible to check whether the planned motion is consistent with the stated high-level decision. 
Although other forms of verifiable language interfaces may also be possible, this work focuses on how far a simple meta-action interface can go when its verifiability is fully exploited.

To this end, we propose \textbf{DriveMA}, a Driving VLA that instantiates meta-action as a simple verifiable language interface between visual inputs and trajectory generation.
DriveMA exploits this interface through three components: trajectory-grounded meta-action annotation for scalable supervision, action-centric pretraining for driving-domain decision learning, and turn-level credit assignment RL for explicit language-action alignment with dense reward and precise credit assignment.
Together, these components strengthen driving-domain decision knowledge while exploiting meta-action verifiability to produce trajectories that better reflect predicted decision, as illustrated in Fig.~\ref{fig:case1}.

Experiments demonstrate that this simple verifiable-interface-centered design is highly effective. 
On the Waymo Open Dataset for Vision-Based End-to-End Driving (WOD-E2E)~\cite{xu2025wod}, DriveMA already achieves a new state of the art with a 2B model, reaching a \textbf{Rater Feedback Score (RFS) of 8.060}, while its 4B version further improves the state of the art to \textbf{8.079}.
On NAVSIM~\cite{dauner2024navsim}, DriveMA obtains competitive closed-loop planning performance among existing Driving VLAs. 
Ablations further show that action-centric pretraining and turn-level credit assignment RL effectively align language decisions with trajectory planning.
Overall, these results show that the key value of DriveMA lies not in introducing meta-actions as a new language interface, but in fully exploiting their verifiability to turn a simple language interface into an effective mechanism for language-action alignment. Our main contributions are summarized as follows:
\begin{itemize}
    \item We propose \textbf{DriveMA}, a Driving VLA framework that instantiates meta-actions as a simple verifiable intermediate language interface and strengthens high-level decision learning through action-centric pretraining.

    \item We introduce a highly data-efficient turn-level credit assignment RL framework for explicit language-action alignment. It exploits meta-action verifiability to construct dense rewards and assigns them precisely across generation tokens.

    \item DriveMA sets a new state of the art on WOD-E2E already with a 2B model, reaching an RFS of \textbf{8.060}, and further pushes the record to \textbf{8.079} with its 4B version. It also obtains competitive closed-loop planning performance on NAVSIM.
\end{itemize}

\section{Related Work}

\subsection{Language Interfaces for Driving VLAs}

Recent language-based driving systems introduce natural language into end-to-end autonomous driving for instruction following, scene understanding, decision grounding, and interpretability~\cite{shao2024lmdrive,sima2024drivelm}. 
Many Driving VLA methods use language reasoning as the intermediate interface between visual inputs and actions~\cite{wang2025hmvlm,li2025recogdrive,rowe2025poutine,zhou2025autovla,zheng2026driveagentr}. 
For example, DriveVLM~\cite{tian2024drivevlm} decomposes driving into scene description, scene analysis, and hierarchical planning, while AutoVLA~\cite{zhou2025autovla} unifies reasoning and action generation in an autoregressive framework with adaptive thinking modes. 
These works show that language can expose useful intermediate states before trajectory generation. 
Meanwhile, NoRD studies a complementary direction by reducing dense reasoning annotations and language overhead for efficient Driving VLAs~\cite{rawal2026nord}.
Our work focuses on the high-level decision interface that often appears near the end of these language-based pipelines, such as driving intentions, maneuvers, or meta-action-like commands. 
DriveMA therefore isolates meta-action as a simple verifiable language interface and studies how to fully exploit it for supervision, reward design, and credit assignment.

\subsection{Language-Action Alignment}

Language-action alignment is a central challenge for applying VLA models to autonomous driving~\cite{Cui_2025, renz2025simlingo,wang2026linkvla,li2025driver1}. 
SimLingo~\cite{renz2025simlingo} aligns language and actions through multi-task training and Action Dreaming, grounding vision-language understanding in closed-loop driving behaviors. 
LinkVLA~\cite{wang2026linkvla} improves language-action coupling by unifying language and action tokens in a shared token space and introducing an action-understanding objective to learn bidirectional trajectory--language mappings. 
Drive-R1~\cite{li2025driver1} mitigates reasoning-planning misalignment mainly through autonomous-driving domain alignment with 3M Visual Question Answering (VQA) and subsequent reasoning supervision. 
Rather than relying on implicit alignment from auxiliary tasks or domain data, DriveMA exploits the verifiability of meta-actions to explicitly measure and encourage consistency between the model's stated high-level decision and its generated trajectory.

\section{DriveMA}

\subsection{Overview}
\begin{wrapfigure}{l}{0.45\linewidth}

    \centering
    \includegraphics[width=\linewidth]{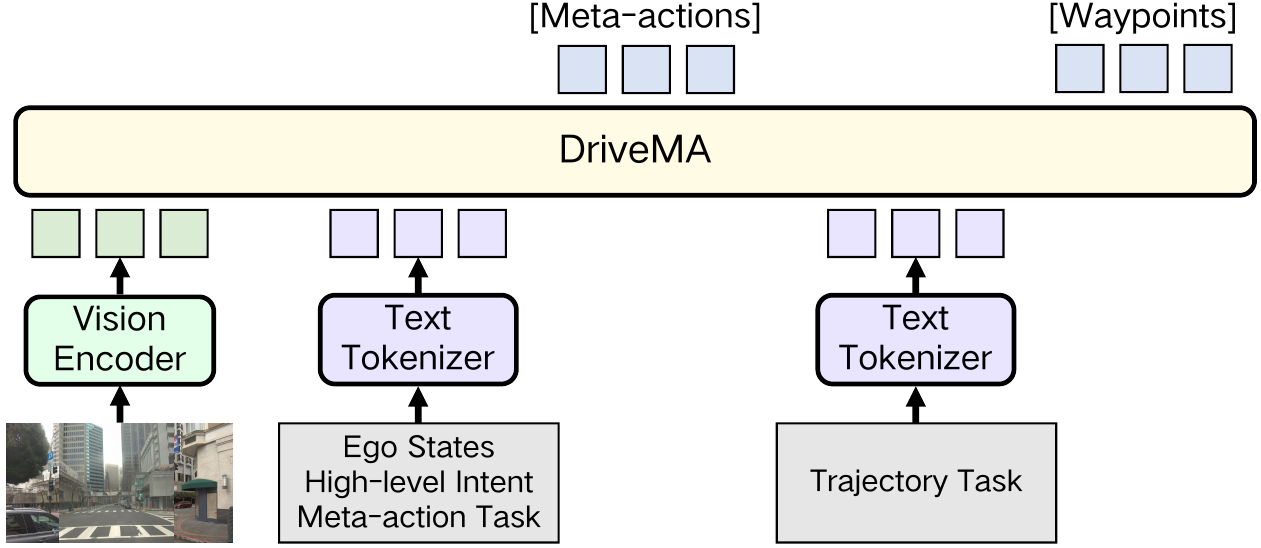}
    \caption{Overview of DriveMA.}
    \label{fig:overview}
\end{wrapfigure}

DriveMA formulates end-to-end driving planning as a meta-action-guided multi-turn generation problem. 
Given a driving input $x$, the model first predicts a compact high-level meta-action $m$, and then generates future waypoints $\tau$ conditioned on both the input and the predicted meta-action:
\[
x \rightarrow m \rightarrow \tau .
\]

As illustrated in Fig.~\ref{fig:overview}, DriveMA builds on a general vision-language model: visual observations are encoded by the vision encoder, while non-visual inputs and outputs, including trajectory coordinates, are represented through the native text tokenizer.

\subsection{Chunk-Based Meta-Action Labeling}


Meta-actions are automatically constructed from expert future trajectories with a trajectory-grounded annotation pipeline.
Given an expert trajectory $\tau^*$ over horizon $T$, we partition it into non-overlapping chunks of size $c$ and annotate each chunk with a discrete meta-action:
\[
m_l^* = \Phi_{\mathrm{ann}}(\tau^*_{[t_l,t_l+c]}).
\]
Each meta-action consists of a longitudinal component, selected from \textit{stop}, \textit{decelerate}, \textit{keep}, and \textit{accelerate}, and a lateral component describing coarse maneuver intent such as \textit{going straight}, \textit{turning}, \textit{lane changing}, or \textit{slight shifting}.
The annotation function $\Phi_{\mathrm{ann}}$ combines trajectory-derived geometric cues with dataset-specific thresholds estimated from training trajectories.
Additional annotation details are provided in Appendix~\ref{app:meta_action_labeling}. 
DriveMA uses $c=T$ by default, where a single one-step meta-action summarizes the entire planning horizon.

\subsection{Meta-Action-Guided Supervised Training}

Before RL, DriveMA uses a two-stage supervised fine-tuning (SFT) pipeline to establish meta-action-guided planning.
The first stage adapts the general VLM to driving-domain action understanding~\cite{zhou2025autovla,li2025driver1,zheng2026driveagentr} and warms up the meta-action prediction task, while the second stage trains the full planning sequence conditioned on expert meta-actions.
This staged design separates action-domain adaptation from low-level trajectory generation.

\paragraph{Action-Centric Pretraining.}
This stage contains two types of supervision. 
First, we use meta-action prediction data automatically generated from expert trajectories, where the model is trained to predict the expert meta-action from the driving input only, i.e., $x \rightarrow m^*$, without generating future waypoints. 
This provides a direct warmup for the intermediate decision interface. 
Second, we incorporate 240K action-related Driving VQA samples from public datasets, selecting samples related to driving intention, action decision, and risk-aware behavior. 
These samples further align the general VLM with driving-domain perception and decision knowledge.

\paragraph{Meta-Action-Conditioned Planning SFT.}
After action alignment pretraining, we train the full meta-action-guided planning task, where each sample follows the multi-turn target sequence $x \rightarrow m^* \rightarrow \tau^*$.
This stage teaches the model to condition trajectory generation on the expert meta-action decision.

\subsection{Turn-Level Credit Assignment Reinforcement Learning}

Although supervised fine-tuning introduces the meta-action language interface into planning and improves the model’s planning ability, it may still produce trajectories that are inconsistent with its own high-level decisions. To address this language-action gap, DriveMA exploits the verifiability of meta-actions and introduces a turn-level credit assignment RL framework. 

\paragraph{Multi-Turn Rollout.}
Given a driving input $x$, we sample $G$ completions from the old policy $\pi_{\theta_{\mathrm{old}}}$, denoted as $o^i=(o_1^i,o_2^i)$ for $i=1,\dots,G$, where $o_1^i$ is the meta-action prediction turn and $o_2^i$ is the trajectory prediction turn. 
Let $T_j^i$ denote the token indices of the $j$-th assistant turn in the $i$-th completion.

\begin{figure}[!htbp]
    \centering
    \includegraphics[width=0.9\linewidth]{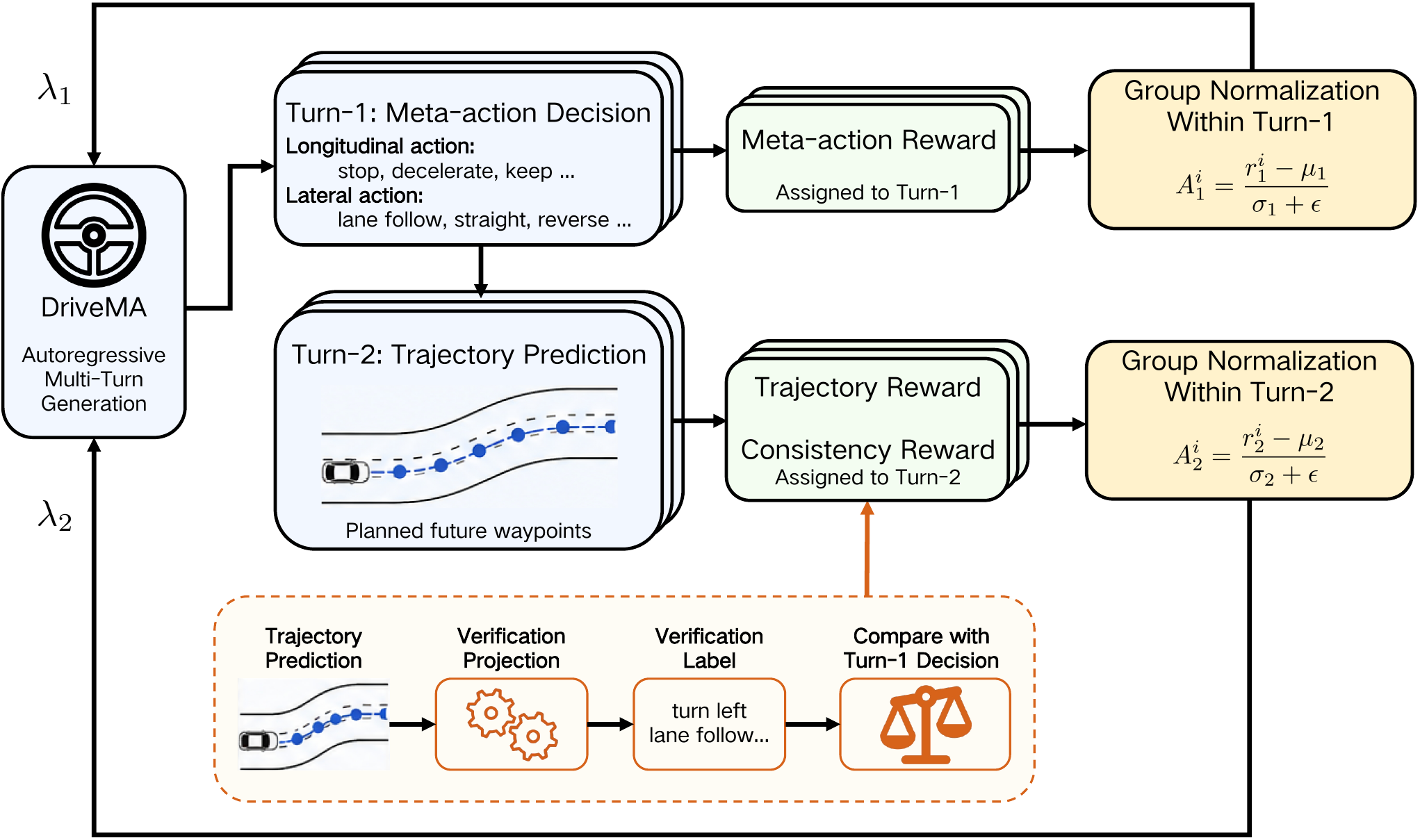}
    \caption{\textbf{Turn-level credit assignment RL in DriveMA.} DriveMA assigns meta-action rewards to the decision turn and trajectory/consistency rewards to the planning turn. Rewards are normalized within each turn, enabling dense supervision and precise credit assignment.}
    \label{fig:RL}
\end{figure}

\paragraph{Dense Reward Design.}
DriveMA assigns separate rewards to the two assistant turns:
\[
r_1^i = R_{\mathrm{meta}}(o_1^i, M^*), \qquad
r_2^i = w_{\mathrm{traj}} R_{\mathrm{traj}}(o_2^i, \tau^*) 
+ w_{\mathrm{cons}} R_{\mathrm{cons}}(o_1^i, o_2^i).
\]
The first reward evaluates meta-action correctness, while the second combines trajectory quality with language-action consistency.
For consistency, we parse the predicted trajectory $\hat{\tau}^i$ from $o_2^i$ and map it to a trajectory-derived verification space with $\Phi_{\mathrm{ver}}$.
The predicted meta-action $\hat{M}^i$ is also projected into this space with $\Gamma(\cdot)$:
\[
R_{\mathrm{cons}}(o_1^i,o_2^i)
=
S\left(\Phi_{\mathrm{ver}}(\hat{\tau}^i), \Gamma(\hat{M}^i)\right),
\]
where $S(\cdot,\cdot)$ measures agreement in the verification space.
This reward encourages the generated trajectory to execute the verifiable motion intent of the predicted meta-action.

\paragraph{Turn-Level Advantage Normalization.}
Standard GRPO~\cite{Guo_2025,shao2024deepseekmath} assigns a single sequence-level reward to all generated tokens, which is suboptimal for our multi-turn setting because different turns optimize different objectives. 
DriveMA instead normalizes rewards separately within each turn:
\[
A_j^i = \frac{r_j^i-\mu_j}{\sigma_j+\epsilon}, \quad j\in\{1,2\},
\]
where $\mu_j$ and $\sigma_j$ are computed over the $G$ sampled completions for the $j$-th turn. 
Thus, meta-action rewards update only meta-action tokens, while trajectory and consistency rewards update trajectory tokens, yielding more precise credit assignment.

\paragraph{Optimization Objective.}
The final turn-level credit assignment GRPO objective is:
\[
\mathcal{J}_{\mathrm{DriveMA}}(\theta)
=
\mathbb{E}_{q,\{o^i\}_{i=1}^{G}\sim\pi_{\theta_{\mathrm{old}}}}
\left[
\frac{1}{G}\sum_{i=1}^{G}
\frac{
\sum_{j=1}^{2}\lambda_j
\frac{1}{|T_j^i|}
\sum_{t\in T_j^i}
s_{j,t}^i(\theta)
}{
\sum_{j=1}^{2}\lambda_j
}
\right]
-
\beta D_{\mathrm{KL}}(\pi_\theta\|\pi_{\mathrm{ref}}),
\]
where $\lambda_j$ controls the relative weight of each turn, and the clipped policy surrogate is
$
s_{j,t}^i(\theta)=
\min\left(
\rho_{i,t}(\theta) A_j^i,\,
\operatorname{clip}(\rho_{i,t}(\theta),1-\epsilon_{\mathrm{low}},1+\epsilon_{\mathrm{high}}) A_j^i
\right),
$
with $\rho_{i,t}(\theta)=\pi_\theta(o_t^i\mid q,o_{<t}^i)/\pi_{\theta_{\mathrm{old}}}(o_t^i\mid q,o_{<t}^i)$. 
Here, \(\epsilon_{\mathrm{low}}\) and \(\epsilon_{\mathrm{high}}\) are the lower and upper clipping bounds.
The reference policy \(\pi_{\mathrm{ref}}\) is the fixed model before RL, and the KL term, weighted by \(\beta\), is applied as a separate regularizer to constrain policy drift.

\section{Experiments}

\subsection{Experiments Setup}

\paragraph{Datasets.}
We evaluate DriveMA on two end-to-end driving benchmarks: WOD-E2E~\cite{xu2025wod} and NAVSIM~\cite{dauner2024navsim}.
WOD-E2E focuses on challenging long-tail scenarios such as construction zones and safety-critical interactions, and we primarily report its official preference-based metric, RFS.
NAVSIM is a planning-oriented autonomous driving benchmark built on OpenScene/nuPlan, and we primarily report PDMS, a closed-loop metric covering safety, progress, and comfort.
In addition, we use open-source driving VQA datasets, including WaymoQA~\cite{waymoqa}, IDKB~\cite{IDKB}, and LingoQA~\cite{marcu2024lingoqa}, comprising a total of 240K samples, for action-centric pretraining.

\paragraph{Implementation Details.}
For Waymo, we downsample the provided 10Hz data by a factor of five for training, and DriveMA predicts 5s future trajectories at 1Hz.
During RL, we use the official preference subset with 479 expert-scored trajectory samples and instantiate $R_{\mathrm{traj}}$ with RFS.
For NAVSIM, DriveMA predicts 4s future trajectories at 1Hz, uses the official trainval split with 103K samples, and holds out 18K samples for RL with $R_{\mathrm{traj}}$ instantiated by PDMS.
For both benchmarks, the model takes only current-frame images from the three front-facing cameras as visual input. 
DriveMA is built on the Qwen3.5 model family~\cite{qwen35blog}, more hyperparameters and prompts are provided in the Appendix~\ref{app:hyperparameters}.

\subsection{Main Results}

\begin{table*}[!h]
\centering
\caption{
Results on the WOD-E2E.
$\uparrow$ indicates higher is better, while $\downarrow$ indicates lower is better.
Best results are shown in \textbf{bold}, and second-best results are \underline{underlined}.
}
\label{tab:waymo_main_results}
\resizebox{0.8\textwidth}{!}{
\begin{tabular}{l c c c c}
\toprule
\textbf{Method} 
& \textbf{RFS Overall $\uparrow$} 
& \textbf{RFS Spotlight $\uparrow$} 
& \textbf{ADE@5s $\downarrow$} 
& \textbf{ADE@3s $\downarrow$} \\
\midrule

\multicolumn{5}{l}{\textit{E2E planning methods}} \\
Swin-Trajectory~\cite{zhou2025swinfrajectory}           & 7.543 & 6.670 & 2.814 & 1.208 \\
DiffusionLTF~\cite{Nguyen2025DiffusionLTF}              & 7.717 & 6.414 & 2.891 & 1.356 \\
UniPlan~\cite{liao2025diffusiondrive}  & 7.780 & 6.654 & 2.986 & 1.308 \\
FROST-Drive~\cite{dong2026frost}           & 7.856 & 7.094 & 3.565 & 2.537 \\
RAP~\cite{feng2026rap}                       & 8.043 & \underline{7.204} & \underline{2.646} & 1.174 \\
\midrule

\multicolumn{5}{l}{\textit{Driving VLA methods}} \\
AutoVLA~\cite{zhou2025autovla}                   & 7.557 & 6.944 & 2.958 & 1.351 \\
NoRD~\cite{rawal2026nord}                      & 7.709 & --     & --     & 1.250 \\
HMVLM~\cite{wang2025hmvlm}                     & 7.737 & 6.727 & 3.072 & 1.327 \\
Poutine~\cite{rowe2025poutine}                   & 7.986 & 6.893 & 2.742 & 1.206 \\
\midrule

\rowcolor{gray!10}
DriveMA-2B                & \underline{8.060} & \textbf{7.251} & \textbf{2.616} & \textbf{1.154} \\
\rowcolor{gray!10}
DriveMA-4B                & \textbf{8.079} & 7.169 & 2.670 & \underline{1.166} \\
\bottomrule
\end{tabular}
}
\end{table*}

\paragraph{WOD-E2E.}
As shown in Table~\ref{tab:waymo_main_results}, DriveMA sets a new state of the art on the WOD-E2E.
DriveMA-2B and DriveMA-4B achieve \textbf{8.060} and \textbf{8.079} RFS Overall, respectively, and outperform all previous methods.
Notably, DriveMA-2B attains the best results on nearly all key metrics, including \textbf{7.251} RFS Spotlight on the official challenging subset and lowest ADE.
This result suggests that even a simple language interface can go surprisingly far when it is verifiable and explicitly optimized for language-action alignment.
Qualitative comparisons with RAP~\cite{feng2026rap} are in Appendix~\ref{app:qualitative_rap}.

\paragraph{NAVSIM Benchmark.}
Table~\ref{tab:navsim_results} reports the closed-loop results on NAVSIM.
DriveMA achieves competitive performance compared with state-of-the-art end-to-end planners, with DriveMA-4B reaching 91.2 PDMS and outperforming existing VLA-based methods.

\begin{table*}[!h]
\centering
\caption{
Results on the NAVSIM v1.
All metrics are reported with one decimal place, and higher is better for all metrics.
Best results are shown in \textbf{bold}, and second-best results are \underline{underlined}.
}
\label{tab:navsim_results}
\footnotesize
\setlength{\tabcolsep}{4.5pt}
\renewcommand{\arraystretch}{1.0}
\resizebox{0.8\textwidth}{!}{
\begin{tabular}{l c c c c c c}
\toprule
\textbf{Method}
& \textbf{PDMS}
& \textbf{Collision}
& \textbf{Area}
& \textbf{Progress}
& \textbf{TTC}
& \textbf{Comfort} \\
\midrule

\multicolumn{7}{l}{\textit{End-to-end planning methods}} \\
TransFuser~\cite{TransFuser}
& 84.0 & 97.7 & 92.8 & 79.2 & 92.8 & \textbf{100.0} \\
Hydra-MDP~\cite{li2024hydra}
& 86.5 & 98.3 & 96.0 & 78.7 & 94.6 & \textbf{100.0} \\
TrajHF~\cite{trajhf}
& 86.4 & 96.6 & 96.6 & 84.5 & 92.1 & \textbf{100.0} \\
DiffusionDriveV2~\cite{DiffusionDriveV2}
& \underline{91.2} & 98.3 & 97.9 & \textbf{87.5} & 94.8 & 99.9 \\
RAP~\cite{feng2026rap}
& \textbf{93.8} & \textbf{99.1} & \underline{98.9} & 86.0 & \underline{96.7} & \textbf{100.0} \\
\midrule

\multicolumn{7}{l}{\textit{Driving VLA methods}} \\
AutoVLA~\cite{zhou2025autovla}
& 89.1 & 98.4 & 95.6 & 81.9 & \textbf{98.0} & 99.9 \\
DriveVLA-W0~\cite{li2025drivevla}
& 90.2 & 98.7 & \textbf{99.1} & 83.3 & 95.3 & 99.3 \\
ReCogDrive~\cite{li2025recogdrive}
& 90.8 & 97.9 & 97.3 & \underline{87.3} & 94.9 & \textbf{100.0} \\
\midrule

\rowcolor{gray!10}
DriveMA-2B
& 90.5 & \underline{98.9} & 97.8 & 85.3 & 95.2 & \textbf{100.0} \\
\rowcolor{gray!10}
DriveMA-4B
& \underline{91.2} & 98.7 & 97.8 & 86.7 & 95.5 & 99.9 \\
\bottomrule
\end{tabular}
}
\end{table*}


\subsection{Ablation Study of Training Strategy}

Table~\ref{tab:training_ablation} analyzes the contribution of each training component.
Compared with direct trajectory SFT, meta-action-guided SFT improves RFS Overall from 7.741 to 7.804 and reduces ADE@5s from 3.065 to 2.802, showing that the meta-action interface provides useful but limited gains by itself.
Action-centric pretraining further improves RFS Overall to 7.893 and RFS Spotlight to 7.189, suggesting better generalization to challenging scenarios.
Turn-level credit assignment further improves over vanilla GRPO under comparable reward supervision, increasing RFS Overall from 7.978 to 8.060 by precisely assigning rewards to their corresponding generation turns.

\begin{table}[!h]
\centering
\caption{
Ablation study of the training strategy.
L-A Consistency is computed in the same way as the language-action consistency reward.
ACP denotes Action-Centric Pretraining.
All RL variants are initialized from Meta-Action SFT w/ ACP.
Vanilla GRPO uses the full reward, but applies them without turn-level credit assignment.
}
\label{tab:training_ablation}
\setlength{\tabcolsep}{6pt}
\renewcommand{\arraystretch}{1.08}
\resizebox{0.85\linewidth}{!}{
\begin{tabular}{lcccc}
\toprule
\textbf{Training Objective}
& \textbf{RFS Overall $\uparrow$}
& \textbf{RFS Spotlight $\uparrow$}
& \textbf{ADE@5s $\downarrow$}
& \textbf{L-A Consistency (\%) $\uparrow$} \\
\midrule

\multicolumn{5}{l}{\textit{Planning interface and SFT baselines}} \\
Direct Trajectory SFT
& 7.741 & 7.094 & 3.065 & -- \\
Meta-Action SFT w/o ACP
& 7.804 & 6.832 & 2.802 & 90.59 \\
Meta-Action SFT w/ ACP
& 7.893 & 7.189 & 2.774 & 91.31 \\
\midrule

\multicolumn{5}{l}{\textit{RL baseline}} \\
Vanilla GRPO w all rewards
& 7.978 & 7.087 & 2.660 & 95.95 \\
\midrule

\multicolumn{5}{l}{\textit{Reward ablation under turn-level credit assignment RL}} \\
$R_{\mathrm{traj}}$
& 8.009 & 7.146 & 2.627 & 88.50 \\
$R_{\mathrm{traj}} + R_{\mathrm{cons}}$
& 8.052 & 7.029 & 2.674 & \textbf{98.80} \\
$R_{\mathrm{traj}} + R_{\mathrm{meta}}$
& 8.038 & 7.018 & 2.646 & 91.32 \\
\rowcolor{gray!10}
$\mathbf{R_{\mathrm{traj}} + R_{\mathrm{cons}} + R_{\mathrm{meta}}}$
& \textbf{8.060} & \textbf{7.251} & \textbf{2.616} & 98.79 \\
\bottomrule
\end{tabular}
}
\end{table}

\begin{figure*}[!h]
    \centering
    \begin{minipage}[t]{\textwidth}
        \centering
        \includegraphics[width=0.9\linewidth]{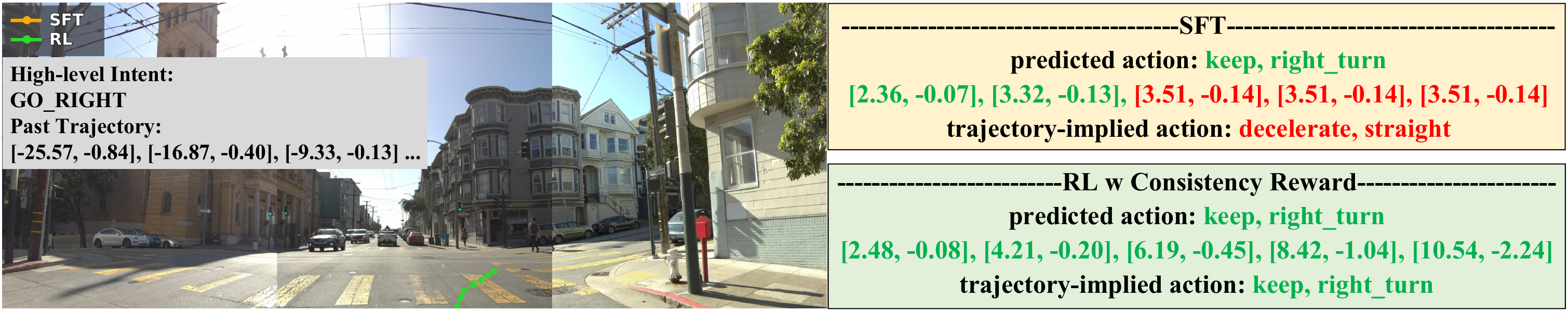}
    \end{minipage}
    \caption{
    \textbf{Qualitative examples of language-action alignment.}
    SFT predicts a high-level meta-action but generates a trajectory with inconsistent longitudinal and lateral behavior.
With the consistency reward, DriveMA generates trajectories that better reflect the predicted meta-action.
    }
    \label{fig:qualitative_alignment}
\end{figure*}

The reward ablation shows that dense language-action supervision is crucial.
Adding $R_{\mathrm{cons}}$ substantially improves L-A Consistency from 88.50\% to 98.80\%, indicating that the consistency reward effectively aligns predicted trajectories with stated meta-actions.
As qualitatively shown in Fig.~\ref{fig:qualitative_alignment}, SFT can predict a right-turn meta-action but generate a trajectory that is nearly straight and prematurely stops.
The consistency reward mitigates this mismatch by making the trajectory execute the predicted decision.
The full reward combination performs best, confirming these objectives are complementary.

\begin{figure}[!h]
    \centering
    \begin{minipage}[t]{0.50\linewidth}
        \centering
        \includegraphics[width=\linewidth]{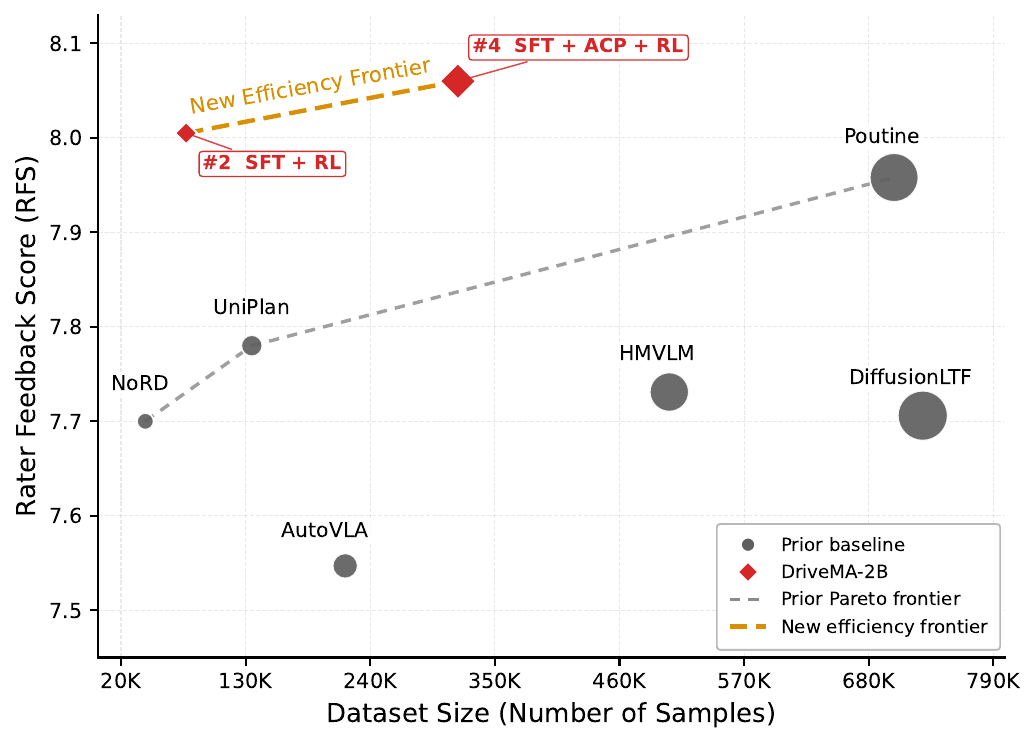}
        \centerline{\small (a) Data efficiency frontier}
    \end{minipage}
    \hfill
    \begin{minipage}[t]{0.4\linewidth}
        \centering
        \includegraphics[width=\linewidth]{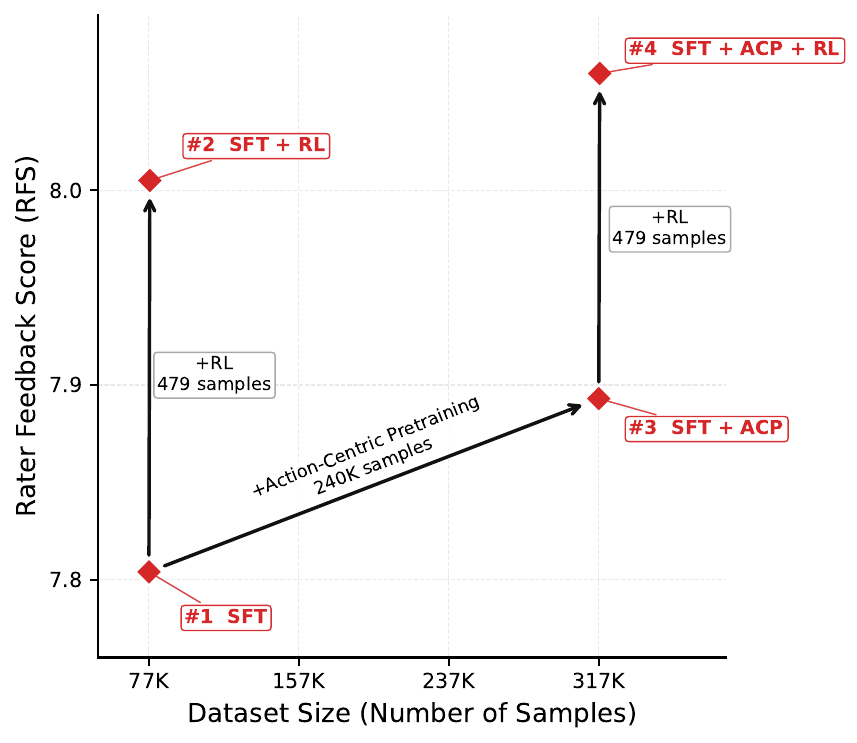}
        \centerline{\small (b) Training component gains}
    \end{minipage}
    \caption{
    \textbf{Data-efficient language-action alignment.}
    ACP denotes Action-Centric Pretraining.
    Left: DriveMA shifts the data-efficiency frontier on WOD-E2E, achieving stronger RFS with substantially fewer training samples than prior methods. Dataset sizes of prior methods are taken from NoRD~\cite{rawal2026nord}.
    Right: the gains from RL and ACP are decomposed over DriveMA-2B variants.
    }
    \label{fig:data_efficiency}
\end{figure}

\subsection{Data-Efficient Language-Action Alignment}
\label{subsec:data_efficiency}

Fig.~\ref{fig:data_efficiency} further analyzes the data efficiency of DriveMA on WOD-E2E.
Compared with existing methods, DriveMA establishes a new efficiency frontier: with only 77K planning samples and 479 preference samples, DriveMA-2B already reaches 8.005 RFS, surpassing prior VLA baselines trained with substantially larger data.
This improvement mainly comes from turn-level credit assignment RL, which effectively exploits limited preference supervision to strengthen language-action alignment.
Notably, adding only 479 RL samples improves direct SFT from 7.804 to 8.005 RFS, showing that verifiable meta-actions enable stable and sample-efficient reward optimization.
Action-Centric Pretraining (ACP) further improves the model by injecting driving-domain action knowledge, and its combination with RL achieves 8.060 RFS using only 77K planning samples, 240K action-centric pretraining samples, and 479 preference samples.
These results suggest that DriveMA does not rely on scaling up supervision alone; instead, verifiable meta-actions allow limited high-quality feedback to be converted into substantial planning gains.

\subsection{Meta-Action Granularity Analysis}

We analyze the temporal granularity of meta-actions on the WOD-E2E validation subset under supervised fine-tuning.
Specifically, we vary the longitudinal chunk size from a single 5s action to finer multi-step actions, while keeping the lateral component as one 5s action.
We report 5s Final Displacement Error (FDE) for 2B, 4B, and 9B models under both predicted and oracle meta-action settings, where oracle meta-actions are replaced with ground-truth labels.

\begin{figure}[t]
\centering
\begin{minipage}[t]{0.63\linewidth}
    \vspace{0pt}
    \centering
    \includegraphics[width=\linewidth]{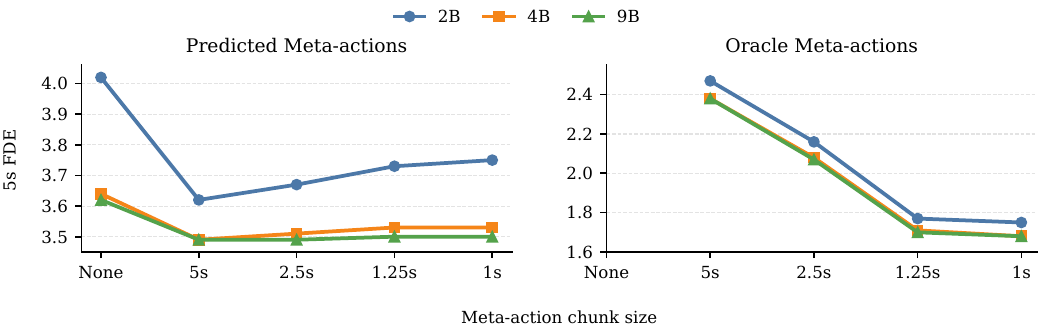}
    \caption{
    Meta-action granularity analysis on the WOD-E2E validation subset.
    The plot reports 5s FDE under predicted and oracle meta-action settings across model scales.
    }
    \label{fig:chunk_size_ablation}
\end{minipage}
\hfill
\begin{minipage}[t]{0.35\linewidth}
    \centering
    \footnotesize
    \setlength{\tabcolsep}{4pt}
    \renewcommand{\arraystretch}{1.08}
    \captionof{table}{
    Complexity of longitudinal action patterns under different chunk sizes in the WOD-E2E training set.
    }
    \label{tab:longitudinal_action_complexity}
    \begin{tabular}{lcc}
    \toprule
    \textbf{Size} & \textbf{Obs.} & \textbf{Entropy} \\
    \midrule
    5s    & 4   & 1.761 \\
    2.5s  & 14  & 2.871 \\
    1.25s & 122 & 4.453 \\
    1s    & 221 & 5.100 \\
    \bottomrule
    \end{tabular}
\end{minipage}
\end{figure}

As shown in Fig.~\ref{fig:chunk_size_ablation}, finer-grained meta-actions improve oracle performance, indicating that they can provide more informative guidance when given correctly.
However, this benefit does not transfer to the predicted setting.
Predicted performance saturates around the 5s one-step meta-action, and scaling from 4B to 9B brings limited additional gain.
This is because finer chunks substantially enlarge the decision space: Table~\ref{tab:longitudinal_action_complexity} shows that observed longitudinal action patterns increase from 4 to 221 as the chunk size decreases from 5s to 1s.
These results suggest that, before RL alignment, the key bottleneck is not only the expressiveness of the meta-action interface but also its predictability.
We therefore use one-step meta-actions as DriveMA's default interface; this analysis also motivates focusing the main evaluation on 2B and 4B variants rather than further backbone scaling.

\section{Conclusion}
\label{sec:conclusion}

In this paper, we show that verifiability is a key property for turning language into an actionable interface in Driving VLAs.
Rather than treating meta-action as a new action vocabulary, DriveMA uses it as a simple instance of a verifiable intermediate language interface.
With trajectory-grounded annotation, action-centric pretraining, and turn-level credit assignment RL, DriveMA explicitly aligns high-level language decisions with low-level trajectory planning.
Despite the simplicity of this interface, DriveMA achieves a new state of the art on WOD-E2E, competitive performance on NAVSIM, and strong data efficiency with limited preference supervision.
These results suggest that future Driving VLAs should not only design more expressive language interfaces, but also make intermediate language states verifiable and optimizable for language-action alignment.

\section{Limitations}
DriveMA currently exploits verifiability through meta-actions, whose consistency reward relies on projecting generated trajectories into a trajectory-derived verification space.
Extending this idea to other language interfaces, such as richer reasoning traces or structured semantic plans, requires interface-specific verification and reward design.
In addition, automatically constructed meta-action labels depend on discretization rules, thresholds, and fine-grained lateral disambiguation, which may introduce noise for ambiguous cases near decision boundaries.
Future work can explore richer yet still verifiable interfaces that preserve the optimization benefits of meta-actions while improving expressiveness.


\clearpage


\bibliography{example}  

\newpage
\appendix

\section{Additional Details on Meta-Action Labeling and Verification}
\label{app:meta_action_labeling}

DriveMA constructs meta-action labels automatically from expert future trajectories.
Each meta-action label consists of a longitudinal component and a lateral component.
The longitudinal component contains four labels: \texttt{stop}, \texttt{decelerate}, \texttt{keep}, and \texttt{accelerate}, where \texttt{stop} also covers waiting behavior.
The lateral component contains ten maneuver labels:
\begin{center}
\small
\begin{tabular}{ll}
\texttt{straight} & \texttt{lane\_follow} \\
\texttt{left\_turn} & \texttt{right\_turn} \\
\texttt{left\_lane\_change} & \texttt{right\_lane\_change} \\
\texttt{left\_shift\_slightly} & \texttt{right\_shift\_slightly} \\
\texttt{reverse} & \texttt{turn\_around}
\end{tabular}
\end{center}

The annotation pipeline is trajectory-grounded.
Longitudinal labels are derived from future trajectory geometry with deterministic rules, while lateral labels are first resolved by trajectory rules whenever possible and use constrained semantic disambiguation only for visually dependent fine-grained cases.
This design provides scalable training labels without human annotation, while preserving a trajectory-derived verification space for consistency reward computation.

\paragraph{Longitudinal annotation.}
For each trajectory chunk, we compute motion statistics including maximum speed, total displacement, path length, and the slope of the speed profile.
Near-static trajectories are identified by clustering low-level motion features over the training set.
The resulting low-motion cluster is converted into deterministic thresholds on maximum speed, total displacement, and path length, yielding the rule for \texttt{stop}.
For the remaining non-static trajectories, we use the speed-profile slope as the acceleration-trend feature.
This feature is clustered into three ordered groups, whose boundaries define \texttt{decelerate}, \texttt{keep}, and \texttt{accelerate}.
After threshold estimation, all longitudinal labels are assigned by fixed deterministic rules.

\begin{table}[h]
\centering
\small
\caption{
Longitudinal Meta-Action annotation rules.
Thresholds are estimated from training trajectories and then fixed for label construction and consistency verification.
Numerical values are rounded to three decimals for readability.
}
\label{tab:longitudinal_rules}
\begin{tabular}{ll}
\toprule
\textbf{Label} & \textbf{Rule} \\
\midrule
\texttt{stop}
& $v_{\max} \le 1.333$, $d_{\mathrm{total}} \le 1.952$, $l_{\mathrm{path}} \le 1.957$ \\
\texttt{decelerate}
& $k_v \le -0.448$ \\
\texttt{keep}
& $-0.448 < k_v < 0.429$ \\
\texttt{accelerate}
& $k_v \ge 0.429$ \\
\bottomrule
\end{tabular}
\end{table}

Here, $v_{\max}$ denotes the maximum speed in meters per second, $d_{\mathrm{total}}$ denotes total displacement in meters, $l_{\mathrm{path}}$ denotes path length in meters, and $k_v$ denotes the slope of the speed profile over time.
A chunk is labeled as \texttt{stop} only when all three low-motion conditions are satisfied.
Otherwise, the label is determined by the acceleration-trend rule.
This gives a clustering-to-rule procedure: clustering discovers dataset-specific motion boundaries, while the final annotation remains deterministic and interpretable.

\paragraph{Lateral annotation.}
The lateral label is obtained with a trajectory-first pipeline.
For each trajectory chunk, we compute geometric features including net heading change, maximum lateral displacement, final lateral displacement, path length, and motion direction.
We denote the net heading change by $\Delta\psi$, the maximum lateral displacement by $d^{\max}_{\mathrm{lat}}$, and the final lateral displacement by $d^{\mathrm{end}}_{\mathrm{lat}}$.

Trajectory rules first identify geometrically clear cases.
Near-straight trajectories are detected using heading and lateral-displacement thresholds.
Within this near-straight group, \texttt{straight}, \texttt{left\_shift\_slightly}, and \texttt{right\_shift\_slightly} are directly assigned according to final lateral displacement.
Thus, slight-shift labels are rule-resolved fine-grained labels, rather than left- or right-oriented semantic groups.
Backward motion is used to detect \texttt{reverse} when the trajectory is not near-static.
For non-straight trajectories, the sign of heading change and final lateral displacement determines left- or right-oriented maneuver groups.
The main trajectory-based lateral rules are summarized in Table~\ref{tab:lateral_rules}.

\begin{table}[h]
\centering
\small
\caption{
Lateral Meta-Action rule groups.
Trajectory rules directly assign near-straight and slight-shift maneuvers, and produce coarse maneuver groups for visually dependent fine-grained cases.
}
\label{tab:lateral_rules}
\begin{tabular}{ll}
\toprule
\textbf{Case} & \textbf{Rule} \\
\midrule
near-straight group
& $|\Delta\psi| \le 15^\circ$ and $d^{\max}_{\mathrm{lat}} < 1.5$ m \\
\texttt{left\_shift\_slightly}
& near-straight group and $d^{\mathrm{end}}_{\mathrm{lat}} > 0.75$ m \\
\texttt{right\_shift\_slightly}
& near-straight group and $d^{\mathrm{end}}_{\mathrm{lat}} < -0.75$ m \\
\texttt{straight}
& near-straight group and $|d^{\mathrm{end}}_{\mathrm{lat}}| \le 0.75$ m \\
left-oriented group
& $\Delta\psi > 15^\circ$ and $d^{\mathrm{end}}_{\mathrm{lat}} > 0$ \\
right-oriented group
& $\Delta\psi < -15^\circ$ and $d^{\mathrm{end}}_{\mathrm{lat}} < 0$ \\
\bottomrule
\end{tabular}
\end{table}

\paragraph{Fine-grained lateral disambiguation.}
Some non-straight lateral maneuvers cannot be reliably distinguished from ego trajectory geometry alone.
For example, a left-oriented trajectory may correspond to \texttt{left\_turn}, \texttt{left\_lane\_change}, or \texttt{lane\_follow}, depending on road topology and lane structure.
Similarly, right-oriented motion may require visual context to distinguish turning from lane changing or lane following.

For these cases, the trajectory rules first generate a small candidate label set.
A constrained visual resolver then selects one label from this candidate set using the front-left, front, and front-right camera images together with structured trajectory information.
This step is used only to refine visually dependent lateral labels for supervised training.
It is not open-ended action generation: the resolver only chooses from rule-provided candidates, which limits semantic drift and keeps the final label grounded in trajectory evidence.

\paragraph{Verification space for consistency reward.}
The fine-grained meta-action labels above are used as supervised training targets.
For consistency reward computation, DriveMA projects both the predicted meta-action and the generated trajectory into a coarser verification space.
This avoids requiring exact agreement on visually dependent fine-grained lateral distinctions, such as turn versus lane change, when such distinctions cannot be reliably inferred from ego trajectory geometry alone.

Formally, let $\Phi_{\mathrm{ver}}(\hat{\tau})$ map a generated trajectory to its trajectory-derived verification label, and let $\Gamma(\hat{M})$ project the predicted meta-action into the same verification space.
The consistency reward compares these two projected labels:
\[
R_{\mathrm{cons}}(o_1,o_2)
=
S\left(\Phi_{\mathrm{ver}}(\hat{\tau}), \Gamma(\hat{M})\right),
\]
where $S(\cdot,\cdot)$ measures agreement in the verification space.

For the longitudinal component, the verification space is the same four-way label space:
\[
\{\texttt{stop}, \texttt{decelerate}, \texttt{keep}, \texttt{accelerate}\}.
\]
For the lateral component, verification is performed in a coarse but rule-aware space:
\[
\begin{aligned}
\texttt{straight},\texttt{lane\_follow} &\rightarrow \texttt{forward},\\
\texttt{left\_shift\_slightly} &\rightarrow \texttt{left\_shift\_slightly},\\
\texttt{right\_shift\_slightly} &\rightarrow \texttt{right\_shift\_slightly},\\
\texttt{left\_turn},\texttt{left\_lane\_change} &\rightarrow \texttt{left},\\
\texttt{right\_turn},\texttt{right\_lane\_change} &\rightarrow \texttt{right},\\
\texttt{reverse} &\rightarrow \texttt{reverse},\\
\texttt{turn\_around} &\rightarrow \texttt{turn\_around}.
\end{aligned}
\]
Thus, a generated trajectory is considered laterally consistent if it matches the corresponding verification group of the predicted meta-action.
For visually dependent labels such as turn versus lane change, consistency is evaluated at the coarse direction level.
For rule-resolved labels such as \texttt{left\_shift\_slightly} and \texttt{right\_shift\_slightly}, consistency is evaluated directly in their own rule-defined categories.
This design avoids requiring exact agreement on visually dependent fine-grained distinctions while still preserving rule-verifiable slight-shift behavior.

Overall, the annotation stage provides semantically meaningful fine-grained meta-action labels, while the verification stage evaluates only the motion properties that can be robustly checked from generated trajectories.
This separation keeps the training interface expressive enough for planning supervision and the consistency reward reliable enough for reinforcement learning.


\section{Training and Inference Hyperparameters}
\label{app:hyperparameters}

\paragraph{Supervised fine-tuning.}
DriveMA uses a two-stage supervised fine-tuning (SFT) pipeline, consisting of Action-Centric Pretraining (ACP) and meta-action-conditioned planning SFT.
Both stages are trained for 1 epoch with a learning rate of $1\times10^{-5}$ and an effective batch size of 64.
We perform full-parameter fine-tuning, including the vision encoder, language model, and multimodal projector.
All SFT experiments are conducted on 8 NVIDIA A800 GPUs.
The maximum image resolution is controlled by \texttt{max\_pixels}, which is set to $512\times512=262{,}144$ pixels.

\paragraph{Reinforcement learning.}
For RL training, we use a learning rate of $1\times10^{-6}$ and train for 600 steps.
The effective batch size is 64, with \texttt{num\_generations} set to 8 and \texttt{steps\_per\_generation} set to 1.
The KL penalty coefficient is set to $\beta=0.4$, and the sampling temperature is 1.0.
The turn-level credit assignment coefficients are set to $\lambda_1=0.2$ and $\lambda_2=0.8$.
For reward weighting, the trajectory reward has weight 1.0, while the consistency reward has weight 0.5.
As in SFT, RL uses full-parameter fine-tuning and sets \texttt{max\_pixels} to $512\times512=262{,}144$.

\paragraph{Inference.}
During inference, we use the same \texttt{max\_pixels} setting as training.
All reported results are obtained with greedy decoding, and we report the top-1 trajectory prediction.

\section{Training/Inference Prompt Format}
\label{app:prompt_format}

DriveMA uses a two-turn dialogue format for meta-action-guided planning.
The first turn predicts the high-level meta-action from visual observations and non-visual driving states.
The second turn predicts the future ego trajectory conditioned on the same context and the predicted meta-action.
The same prompt template is used during training and inference, except that the target response is only provided during training

\begin{promptbox}[title={Two-turn prompt format for DriveMA}]
Human:
You are an expert driver.

Input:
- 1 frame of multi-view images collected from the ego-vehicle at the present timestep:
  front_left_view: <image>; front_view: <image>;
  front_right_view: <image>
- Current high-level intent: <intent>
- Current acceleration: <current_acceleration>
- Current velocity: <current_velocity>
- 4-second past trajectory (16 steps at 4 Hz): <past_trajectory>
- 4-second past acceleration (16 steps at 4 Hz): <past_acceleration>
- 4-second past velocity (16 steps at 4 Hz): <past_velocity>

Coordinate System Definition:
X-axis: positive forward, negative backward;
Y-axis: positive left, negative right.

Task: Inspect the input and make the decision.
Output format:
longitudinal action: xx, lateral action: xx

Assistant:
longitudinal action: <longitudinal_action>, lateral action: <lateral_action>

Human:
Task: Given the above information, predict the optimal 5-second future trajectory (5 steps at 1 Hz) of the ego vehicle.
Output format:
[x_1, y_1], [x_2, y_2], [x_3, y_3], [x_4, y_4], [x_5, y_5]

Assistant:
[<x_1>, <y_1>], [<x_2>, <y_2>], [<x_3>, <y_3>], [<x_4>, <y_4>], [<x_5>, <y_5>]
\end{promptbox}

\begin{figure*}[!h]
    \centering
    \includegraphics[width=\linewidth]{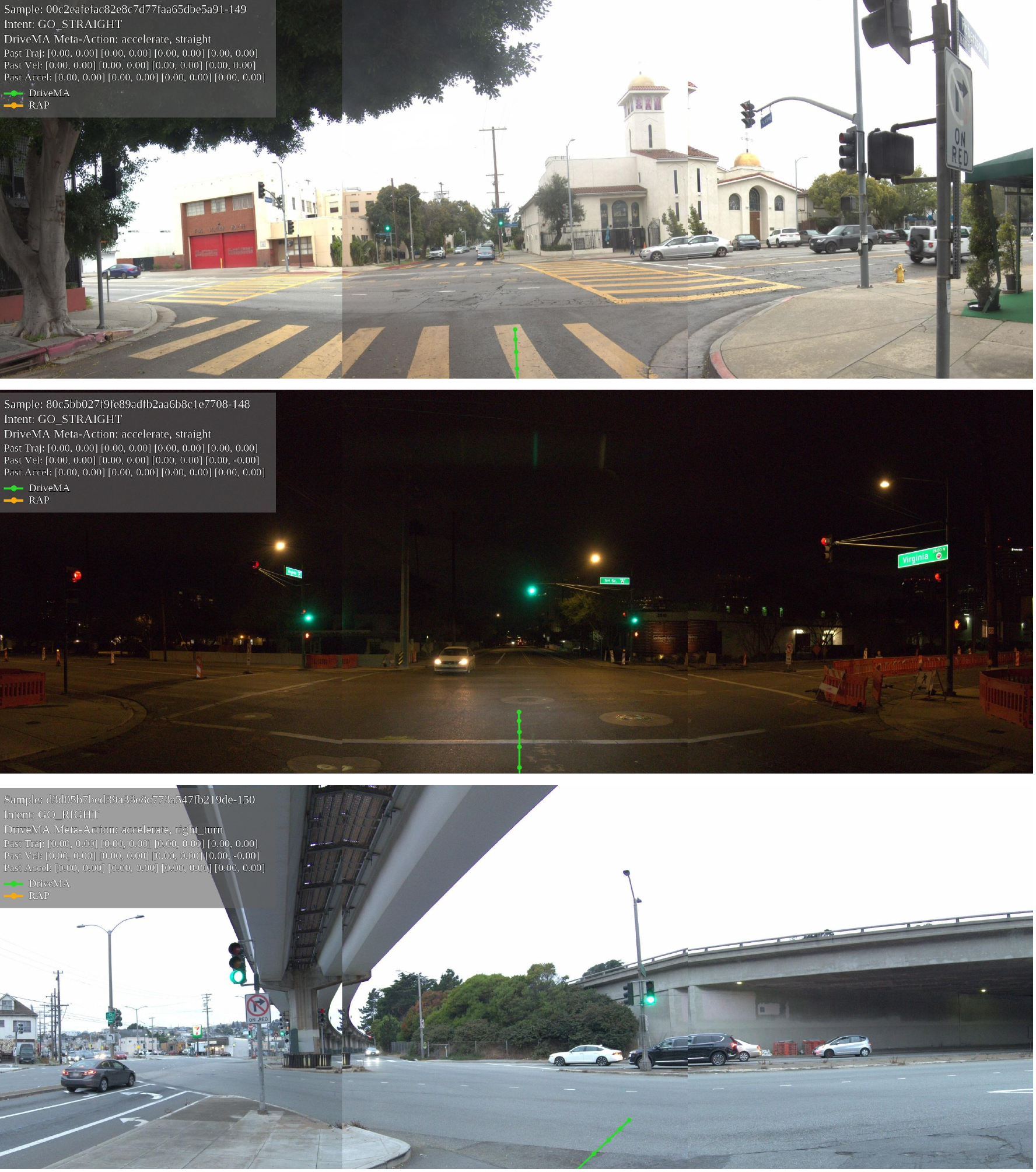}
    \caption{
    \textbf{Qualitative comparison under static-history green-light scenarios.}
    We compare DriveMA with RAP in three cases where the ego vehicle has nearly zero past motion, but the current traffic light indicates that it should start moving.
    DriveMA predicts the meta-action \textit{accelerate} and generates a moving trajectory, while RAP tends to remain close to the static history.
    These cases suggest that the meta-action interface can help reduce history-dominated planning when the future driving intent changes from the recent motion pattern.
    }
    \label{fig:static_history_green_light}
\end{figure*}

\section{Qualitative Comparison with the Previous State of the Art}
\label{app:qualitative_rap}

We further provide qualitative comparisons between DriveMA and RAP~\cite{feng2026rap}, the previous state-of-the-art method on WOD-E2E.
These visualizations are intended to complement the quantitative results by showing representative planning behaviors in challenging scenarios.
In each example, we overlay the predicted trajectories of DriveMA and RAP under the same input scene.
The comparisons highlight cases where DriveMA's predicted meta-action provides an explicit high-level motion intent that is faithfully reflected in the generated trajectory.

\paragraph{Starting from static history at green lights.}
Fig.~\ref{fig:static_history_green_light} shows three representative cases where the ego vehicle has an almost fully static history, while the current traffic light indicates that the vehicle should start moving.
This setting is challenging because a planner can easily overfit to the near-zero past motion and extrapolate a stationary future.
Across all three cases, DriveMA predicts the meta-action \textit{accelerate} and generates a moving trajectory consistent with the green-light context.
In contrast, RAP tends to keep the vehicle nearly stationary, suggesting an over-reliance on historical motion states.
These examples qualitatively support the motivation in Fig.~\ref{fig:case1}: introducing a verifiable meta-action interface helps DriveMA convert language-level driving intent into executable trajectory planning, especially when the desired future motion differs from the recent motion history.

\begin{figure*}[!h]
    \centering
    \includegraphics[width=\linewidth]{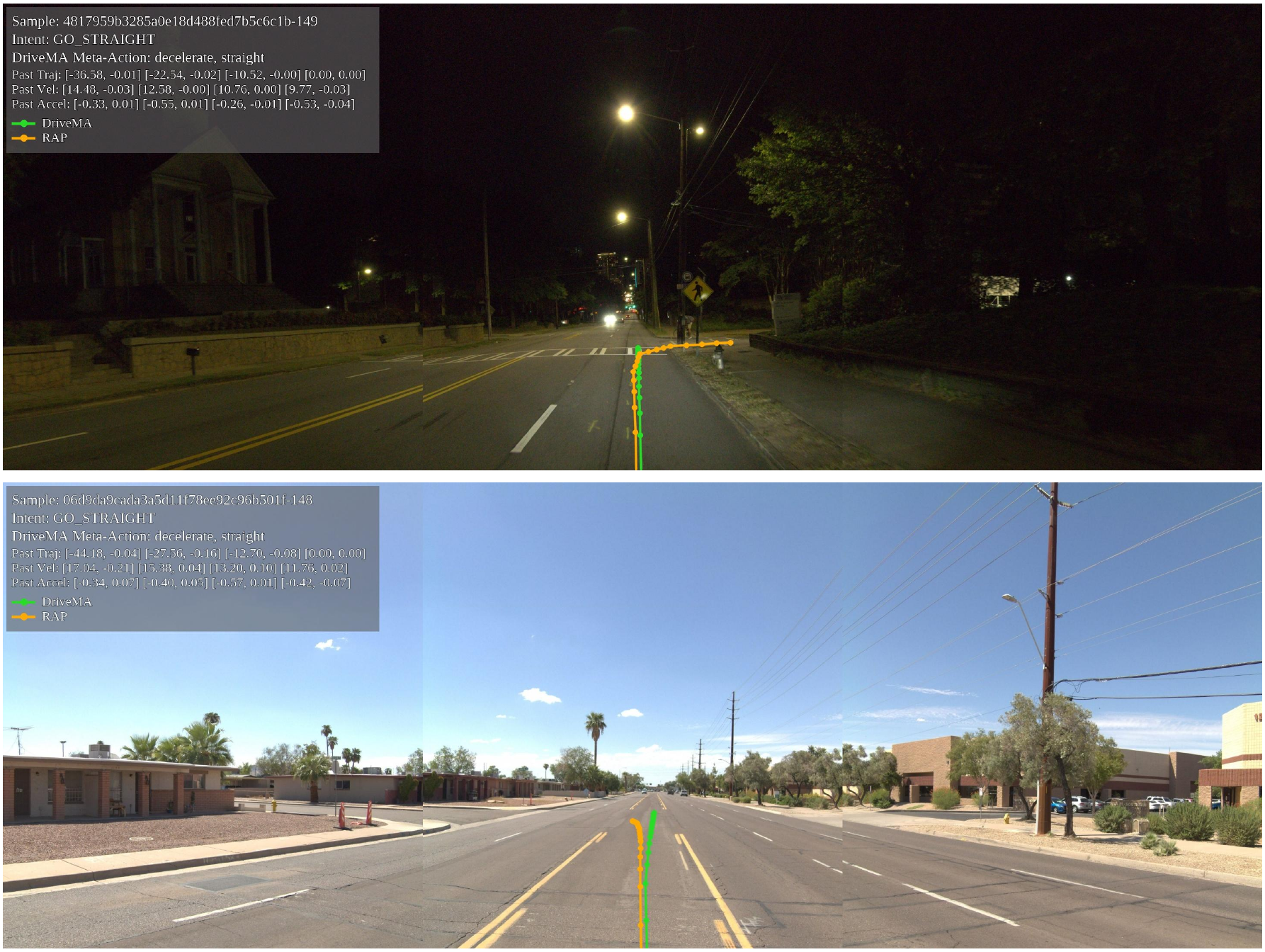}
    \caption{
    \textbf{Qualitative comparison on route-intent consistency.}
    We compare DriveMA with RAP in scenarios where the input intent is \textit{GO\_STRAIGHT}, but nearby side roads provide plausible turning opportunities.
    RAP tends to deviate toward the side road, resulting in trajectories inconsistent with the route-level intent.
    DriveMA predicts the meta-action \textit{decelerate, straight} and generates trajectories that remain aligned with the intended straight-ahead route.
    }
    \label{fig:intent_consistency_side_intersection}
\end{figure*}

\paragraph{Route-intent consistency at side intersections.}
Fig.~\ref{fig:intent_consistency_side_intersection} shows cases where the high-level intent is \textit{GO\_STRAIGHT}, while the scene contains nearby side roads or turning opportunities.
Such scenarios can be ambiguous for trajectory prediction because local road geometry may suggest a feasible turn even when the route-level intent requires continuing straight.
In both examples, RAP generates trajectories that deviate toward the side road, producing a plan inconsistent with the input intent.
In contrast, DriveMA predicts the meta-action \textit{decelerate, straight} and keeps the planned trajectory aligned with the through lane.
In the first case, DriveMA also slows down near the right-side intersection, producing a conservative straight-ahead plan instead of turning toward the side road.
In the second case, DriveMA continues along the current road despite the visible left-side branch.
These examples suggest that the explicit meta-action interface helps DriveMA bind route-level intent to trajectory generation, reducing spurious turning behaviors caused by local intersection geometry.

\begin{figure*}[!h]
    \centering
    \includegraphics[width=\linewidth]{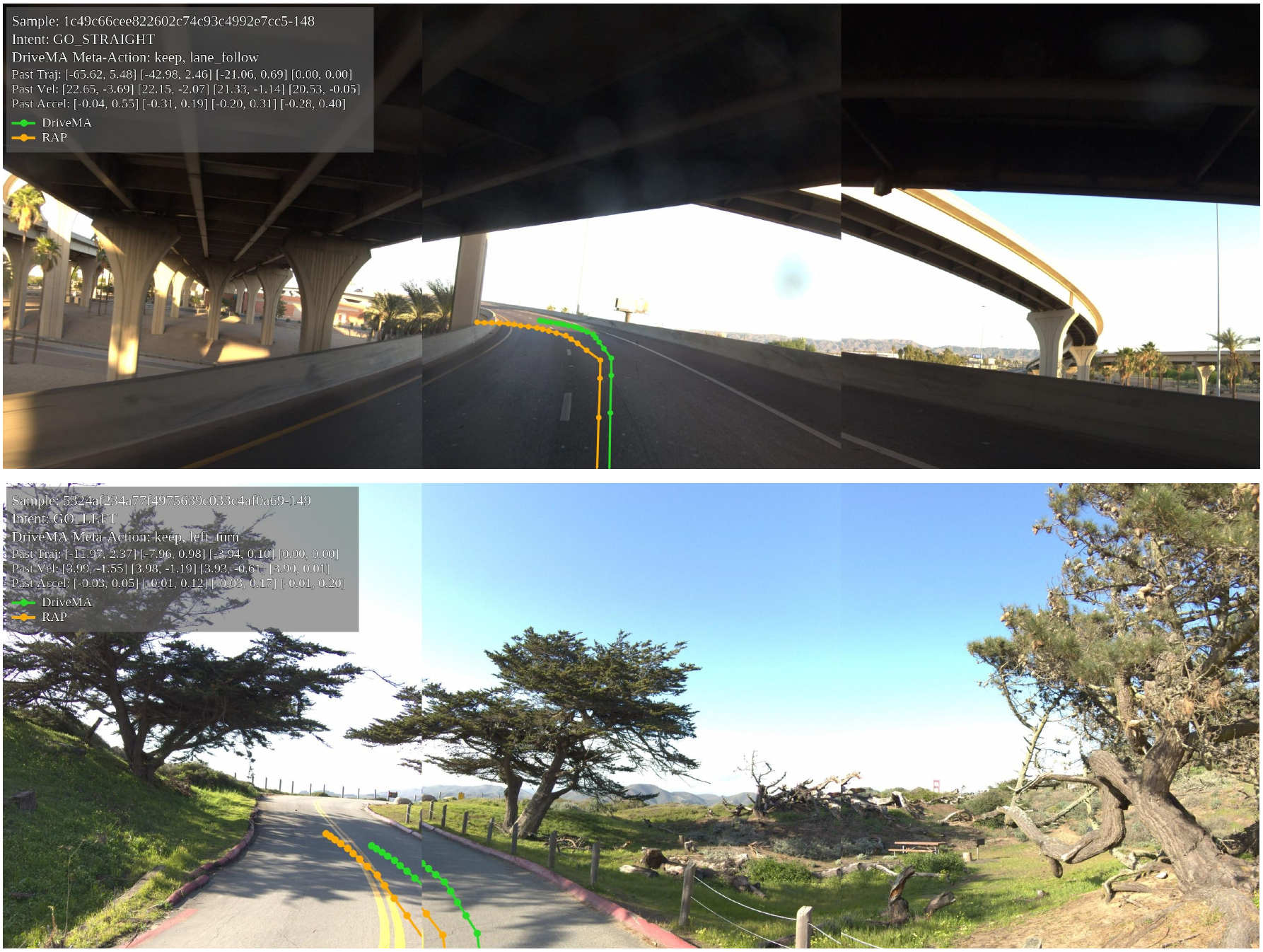}
    \caption{
    \textbf{Qualitative comparison on trajectory curvature control.}
    We compare DriveMA with RAP in curved-road scenarios where the planner must accurately control turning magnitude to remain within the drivable corridor.
    RAP produces overly aggressive turning trajectories, leading to potential road-boundary violation in the first case and lane-center crossing in the second case.
    DriveMA generates smoother trajectories that better follow the lane geometry.
    }
    \label{fig:curvature_control}
\end{figure*}

\paragraph{Trajectory curvature control on curved roads.}
Fig.~\ref{fig:curvature_control} shows two cases where accurate control of trajectory curvature is important for staying within the drivable corridor.
Although the route intent and coarse maneuver are not ambiguous, RAP predicts overly aggressive turning trajectories.
In the first case, RAP bends too sharply toward the road boundary under the overpass, which may lead to unsafe interaction with the curb or roadside structure.
In the second case, RAP cuts across the lane center and approaches the opposing lane on a left-curving road.
In contrast, DriveMA generates smoother trajectories that better follow the lane geometry: it keeps the vehicle within the current lane in the lane-following case and produces a more moderate left-turn trajectory in the curved-road case.
These examples indicate that DriveMA's improvement is not limited to selecting the correct high-level intent; the meta-action-guided planning process also helps produce geometrically better calibrated trajectories under curved road layouts.

\begin{figure*}[!h]
    \centering
    \includegraphics[width=\linewidth]{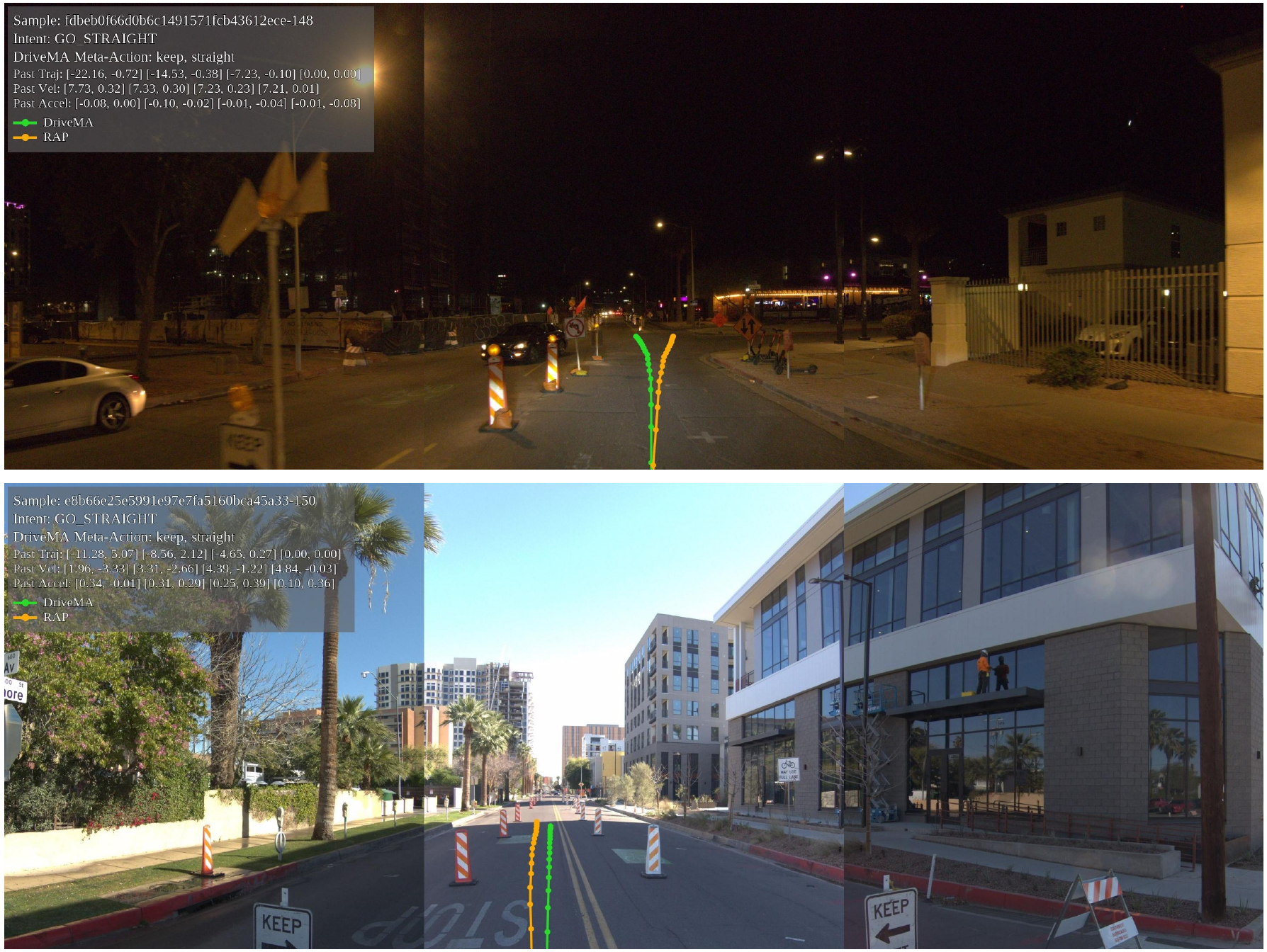}
    \caption{
    \textbf{Qualitative comparison under cone-defined temporary corridors.}
    We compare DriveMA with RAP in construction-zone scenarios where traffic cones reshape the local drivable corridor.
    DriveMA generates trajectories that better follow the temporary corridor, while RAP tends to drift toward the roadside or the cone boundary.
    These cases illustrate DriveMA's ability to adapt planning to local construction constraints.
    }
    \label{fig:cone_corridor}
\end{figure*}

\paragraph{Planning under cone-defined temporary corridors.}
Fig.~\ref{fig:cone_corridor} shows two construction-zone cases where traffic cones define a temporary drivable corridor that differs from the nominal lane geometry.
Such scenarios are challenging because blindly following the original lane marking may lead to unsafe interaction with temporary obstacles, while excessive deviation may violate the intended route.
In the first case, DriveMA keeps a straight trajectory with a slight left adjustment, following the corridor formed by cones on both sides, whereas RAP drifts toward the roadside.
In the second case, the cone-defined corridor is shifted near the double-yellow line.
Although regular driving would normally avoid crossing the double yellow line, DriveMA produces a more centered trajectory within the temporary corridor, while RAP stays closer to the left-side cones and may risk interacting with them.
These examples suggest that DriveMA can better adapt its trajectory to local construction constraints rather than rigidly following nominal lane markings or drifting toward nearby obstacles.

\begin{figure*}[!h]
    \centering
    \includegraphics[width=\linewidth]{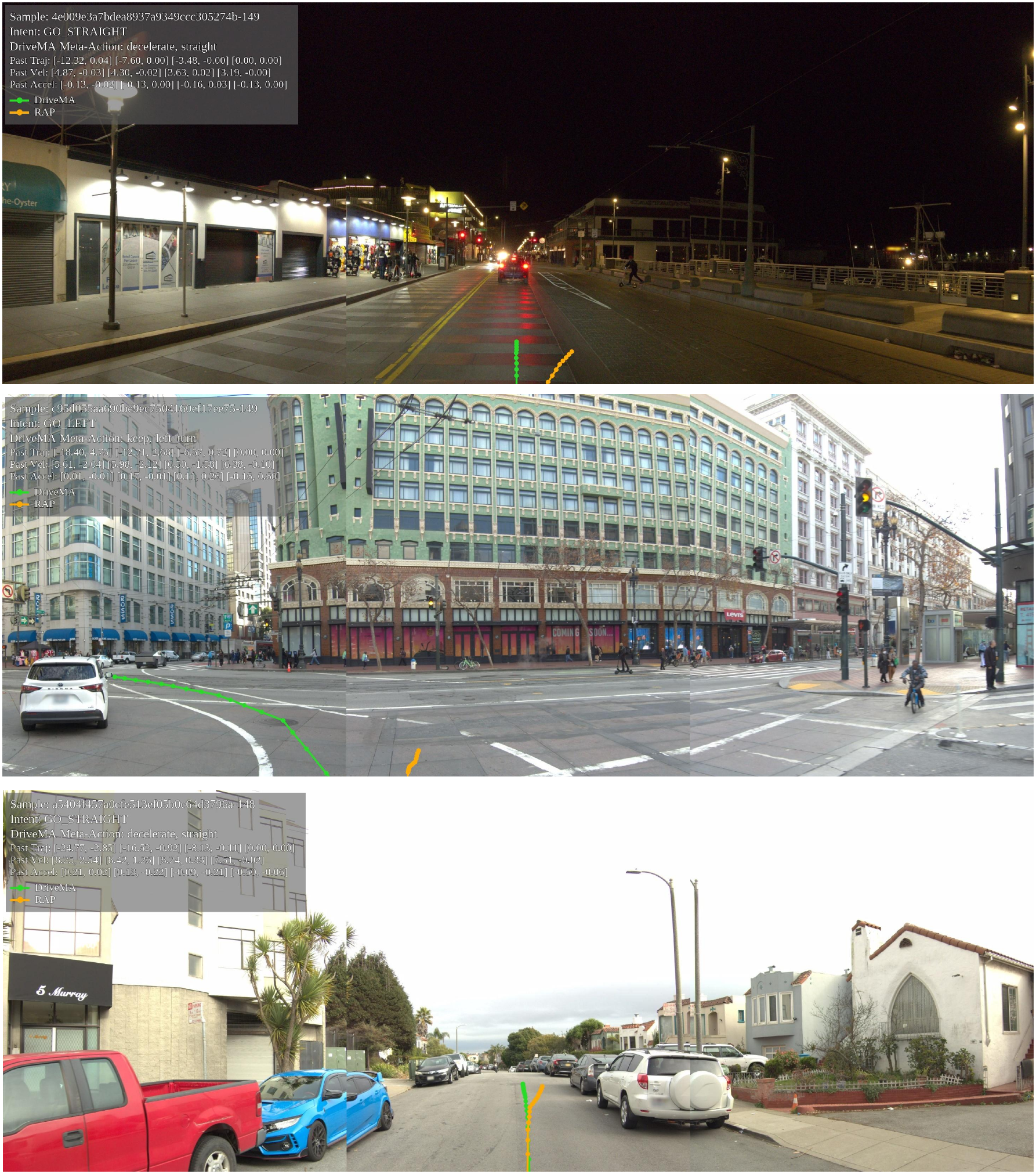}
    \caption{
    \textbf{Qualitative comparison on maneuver-direction errors.}
    We compare DriveMA with RAP in complex scenes involving traffic lights, intersections, and interactive road users.
    RAP sometimes predicts trajectories with incorrect or over-reactive maneuver directions, such as drifting toward the roadside or failing to follow the intended turn.
    DriveMA predicts meta-actions consistent with the route intent and generates trajectories that better preserve the intended driving direction under these scene constraints.
    }
    \label{fig:maneuver_direction_errors}
\end{figure*}

\paragraph{Maneuver-direction errors in complex scenes.}
Fig.~\ref{fig:maneuver_direction_errors} further shows cases where RAP produces trajectories with incorrect maneuver directions under complex scene constraints.
These errors are more severe than small geometric deviations, as they directly conflict with the route intent or the locally feasible driving corridor.
In the first case, the input intent is \textit{GO\_STRAIGHT} and the road ahead is controlled by a red light, where DriveMA predicts \textit{decelerate, straight} and plans a conservative straight trajectory.
RAP instead deviates to the right, although the scene contains left-guidance markings and no clear rightward route.
In the second case, the route intent is \textit{GO\_LEFT}; DriveMA predicts a left-turn meta-action and generates a consistent left-turn trajectory, while RAP fails to follow the intended turning direction.
In the last case, DriveMA decelerates and keeps straight to handle an oncoming road user, while RAP deviates excessively toward the roadside, likely due to over-conservative obstacle avoidance.
These examples suggest that DriveMA is less prone to direction-level planning errors when route intent and local scene constraints must be jointly considered.

\end{document}